\documentclass[conference]{IEEEtran}
\IEEEoverridecommandlockouts
\usepackage{cite}
\usepackage{times}
\usepackage{epsfig}
\usepackage{graphicx}
\usepackage{amsmath}
\usepackage{amssymb}

\usepackage[linesnumbered, ruled, vlined]{algorithm2e}
\usepackage{booktabs}
\usepackage{mathtools}
\usepackage{multicol}
\usepackage{multirow}
\usepackage{bbm}
\usepackage{bm}
\usepackage{wrapfig}
\usepackage{enumitem}
\usepackage{tabularx}
\usepackage{url}
\usepackage[english]{babel}
\usepackage{amsthm}
\newtheorem{theorem}{Theorem}
\newcolumntype{Y}{>{\centering\arraybackslash}X}
\def\BibTeX{{\rm B\kern-.05em{\sc i\kern-.025em b}\kern-.08em
    T\kern-.1667em\lower.7ex\hbox{E}\kern-.125emX}}
\begin{document}

\title{Towards Cross-Domain Continual Learning
\thanks{This research is supported by the Agency for Science, Technology and Research (A*STAR) under its RIE 2025 – Industry Alignment Fund – Pre Positioning
 (IAF-PP) funding scheme (Project No: M23L4a0001).}
}

\author{
\IEEEauthorblockN{1\textsuperscript{st} Marcus de Carvalho}
\IEEEauthorblockA{
\textit{Nanyang Technological University} \\
50 Nanyang Ave, 639798, Singapore \\
{\tt\small marcusvi001@e.ntu.edu.sg}
}
\and
\IEEEauthorblockN{2\textsuperscript{nd} Mahardhika Pratama}
\IEEEauthorblockA{
\textit{University of South Australia} \\
Mawson Lakes, Adelaide, 5095, Australia \\
{\tt\small dhika.pratama@unisa.edu.au}
}
\and
\IEEEauthorblockN{3\textsuperscript{rd} Jie Zhang}
\IEEEauthorblockA{
\textit{Nanyang Technological University} \\
50 Nanyang Ave, 639798, Singapore \\
{\tt\small zhangj@ntu.edu.sg}
}
\and
\IEEEauthorblockN{4\textsuperscript{th} Chua Haoyan}
\IEEEauthorblockA{
\textit{Nanyang Technological University} \\
50 Nanyang Ave, 639798, Singapore \\
{\tt\small haoyan001@e.ntu.edu.sg}
}
\and
\IEEEauthorblockN{5\textsuperscript{th} Edward Yapp}
\IEEEauthorblockA{
\textit{Singapore Institute of Manufacturing Technology (SIMTech)} \\ \textit{Agency for Science, Technology and Research (A*STAR)} \\
5 CleanTechLoop \#01-01, CleanTechTwo Blk B, 636732, Singapore \\
{\tt\small edward\textunderscore yapp@simtech.a-star.edu.sg}
}
}


\maketitle

\begin{abstract}
Continual learning is a process that involves training learning agents to sequentially master a stream of tasks or classes without revisiting past data. The challenge lies in leveraging previously acquired knowledge to learn new tasks efficiently, while avoiding catastrophic forgetting. Existing methods primarily focus on single domains, restricting their applicability to specific problems.

In this work, we introduce a novel approach called Cross-Domain Continual Learning (CDCL) that addresses the limitations of being limited to single supervised domains. Our method combines inter- and intra-task cross-attention mechanisms within a compact convolutional network. This integration enables the model to maintain alignment with features from previous tasks, thereby delaying the data drift that may occur between tasks, while performing unsupervised cross-domain (UDA) between related domains. By leveraging an intra-task-specific pseudo-labeling method, we ensure accurate input pairs for both labeled and unlabeled samples, enhancing the learning process. To validate our approach, we conduct extensive experiments on public UDA datasets, showcasing its positive performance on cross-domain continual learning challenges. Additionally, our work introduces incremental ideas that contribute to the advancement of this field.

We make our code and models available to encourage further exploration and reproduction of our results: \url{https://github.com/Ivsucram/CDCL}
\end{abstract}


\section{Introduction}
\label{sec:intro}

\IEEEPARstart{C}{ontinual}
learning (CL) has gained significant attention in the context of data streaming systems, particularly for self-learning agents and robots. While offline neural networks approaches are typically designed around data collected in controlled environments, aggregated into static datasets, such strategies deviate significantly from the more neural learning conditions experienced by humans and other mammals throughout their lifespans \cite{BabyToRobot}. 

In stark contrast to current dominant artificial learning status quo, human learning involves accumulating and retaining past experiences, seamlessly integrating them when acquiring new skills and tackling novel challenges. The human brain excels at integrating multisensory information, facilitating effective adaptation to uncertain sensory scenarios and leveraging knowledge from complementary domains to accomplish common tasks \cite{Multisensory2014-2}. However, deep learning models, while showing promise on static data, often suffer from {\it catastrophic forgetting}, a consequence of a weak stability-plasticity trade-off \cite{ART}. This phenomenon results in new information overwriting past experiences, leading to a significant decline in performance on previous tasks. Tackling this challenge in CL has been a long-standing pursuit in the fields of machine learning, neural networks, and artificial intelligence \cite{HASSABIS2017245, THRUN199525}.

Despite extensive research on unsupervised domain adaptation \cite{CDTrans, MMDE, decarvalho2021acdc, CrDoCo, li2021cross}, its practical application in CL remains largely unexplored. Current approaches in this field predominantly focus on either single-domain continual learning \cite{A-GEM, GEM, DER, HAL, LVT} or supervised multi-domain continual learning \cite{Simon_2022_CVPR}, both of which have limitations in fully harnessing the wealth of available world knowledge. The challenge lies in addressing two fundamental issues faced by unsupervised domain adaptation methods: task drift, which involves adapting to changes in the conditional distributions of given labels, and domain drift, concerning changes in the marginal distributions of given domains. Consequently, single-domain and supervised multi-domain continual learning methods are susceptible to {\it feature-alignment catastrophic forgetting} when applied to unsupervised multi-domain problems. This vulnerability results in not only overwriting past experiences but also compromising the understanding of domain invariance.

To overcome these challenges, more comprehensive approaches are required to effectively handle both task and domain drifts, ensuring robust performance across diverse domains while preserving acquired knowledge for continual learning scenarios. A compelling real-world example would be a hypothetical self-driving car that initially learns to navigate the streets of London using supervised datasets and then expands that knowledge to unsupervised roads in the countryside. Subsequently, the same car needs to adapt and learn how to drive in France, facing new laws, road signs, landmarks, and driving on the opposite side of the road. The car must maintain its previous knowledge from London and extend it to unsupervised related domains, such as mountain roads and countryside paths, to achieve successful and safe navigation in this new setting. 

In this research, we present a pioneering framework called Cross-Domain Continual Learning (CDCL), aimed at addressing the challenges of unsupervised domain adaptation in the continual learning setting. CDCL introduces a unique mechanism known as {\it inter intra-task cross-attention}, facilitating category-level feature alignment between labeled source domains and unlabeled target domains within the continual learning context. Additionally, CDCL incorporates an innovative {\it intra-task center-aware pseudo-label} procedure, strategically organizing similar source and target domain samples while reducing noise interference. This procedure contributes to more effective learning in a continual setup. Moreover, CDCL leverages a rehearsal memory that not only stores previous source and target data samples but also retains previously generated logits. By doing so, the model is compelled to mimic acquired knowledge when learning new tasks, leading to improved performance and knowledge retention during continual learning processes.

The main contributions of this paper are four-fold:
\begin{itemize}
    \item A novel framework for Cross-Domain Continual Learning (CDCL) to solve the unsupervised cross-domain (UDA) task-incremental learning problem, while maintaining a good stability-plasticity trade-off.
    \item An {\it inter- intra-task cross-attention} mechanism that aligns domain features in current tasks and consolidates previous alignment knowledge, alleviating the feature-alignment catastrophic forgetting of related knowledge domains.
    \item An {\it intra-task-based center-aware pseudo-labeling} method that identifies similar task-specific samples between domains.
    \item An extensive experimental analysis showcasing the success of applying the cross-attention mechanism in the cross-domain task-incremental learning problem, while providing hypotheses towards incremental contributions towards cross-domain class-incremental learning solutions.
\end{itemize}


\section{Related work}
\label{sec:related-work}

\subsection{Unsupervised Domain-Adaptation}
\label{subsec:unsupervised-domain-adptation}

Domain adaptation seeks to create a knowledge representation in which an algorithm cannot discern the origin of input data from either the source or target domain. Specifically, unsupervised domain adaptation (UDA) imposes an additional constraint, wherein the target domain possesses limited or scarce labeled data.

UDA methods currently exist in two main categories. The first is domain-level UDA \cite{MMDE, decarvalho2021acdc, CrDoCo}, which aims to create a latent space that is invariant to domain-specific characteristics. On the other hand, category-level UDA \cite{CDTrans, li2021cross} focuses on generating a feature-aligned network between the two domains, ensuring that their category distributions are closely situated in the latent space. It is worth noting that category-level UDA methods often outperform their domain-level counterparts, given that the resulting domain-invariant latent space allows the categories from both domains to be represented in close proximity. This proximity facilitates effective knowledge transfer and adaptation between the domains, leading to enhanced performance in the category-level UDA methods.

While many computational vision UDA methods still rely on convolutional neural networks (CNN) architecture, vision transformers (ViT) \cite{ViT} have recently emerged as a cutting-edge driver for UDA \cite{CDTrans, DoT}. Despite their state-of-the-art performance, one drawback of the ViT architecture is its substantial data requirements. Consequently, researchers have made several efforts to develop data-efficient visual transformers. Among these endeavors, the most popular ones explicitly incorporate convolutional approaches into the transformer architecture \cite{CCT, dai2021coatnet, Graham_2021_ICCV, HaloNets}. Of particular interest is the CCT (Convolutional-Contrastive Training) method \cite{CCT}, which addresses the need for a class token and positional embedding by employing a sequence pooling strategy and convolutional operations. This innovation leads to more efficient data utilization, reducing the data burden for ViT-based UDA approaches. By integrating convolutional techniques into vision transformers, researchers aim to strike a balance between data efficiency and performance, paving the way for more accessible and effective UDA solutions in the realm of computational vision.

Pseudo-labeling \cite{LEOPARD, PseudoLabel} has emerged as a recent catalyst for enhancing the accuracy of UDA procedures. This approach involves generating synthetic labels for samples, effectively incorporating confident unlabeled data into the training set. In the context of domain adaptation, pseudo-labeling techniques typically leverage the source domain data or a classifier pre-trained on a related domain, such as ImageNet \cite{ImageNet}, to serve as the pseudo-label generator for the target domain data. Several methodologies have been proposed to leverage pseudo-labeling for domain adaptation, such as: Separating sparse feature representations \cite{ATPL} - this technique aims to disentangle the feature representations between the source and target domains, enabling better alignment and adaptation of the model to the target domain; Incorporating regularization terms \cite{Choi2022PseudoLabel} - by adding regularization terms to the loss function, pseudo-labeling encourages the model to generalize better across domains and reduce overfitting; Fine-tuning \cite{CA-UDA} - After pseudo-labels are assigned to the target domain data, fine-tuning the model using this augmented dataset allows for domain-specific adaptation, leading to improved performance on the target domain.
These advancements in pseudo-labeling have shown promise in bridging the domain gap and achieving more accurate and robust UDA solutions. By harnessing synthetic labels and exploiting the relationship between different domains, pseudo-labeling contributes to the effective transfer of knowledge from the source to the target domain, ultimately leading to improved performance and generalization in UDA tasks.

In order to mitigate the adverse impact of noisy labeling, often referred to as miss-synthetic labeling, various approaches have been proposed in the literature. Some works explicitly focus on re-evaluating and correcting erroneous labels \cite{pmlr-v162-liu22w}, while others concentrate on improving the selection of unlabeled samples using the confidence of the model \cite{Nishi_2021_CVPR, Liang2020SHOT}. Additionally, some methods introduce objective functions that exhibit tolerance to labeling noise \cite{SCE}. However, these approaches were primarily designed for static environments and do not address the challenges of continual learning, where a model faces sequential tasks from source and target domains. Consequently, they are susceptible to the issue of catastrophic forgetting in such dynamic scenarios.

To address the limitations of existing methods and tackle cross-domain continual learning problems, CDCL (Cross-Domain Continual Learning) introduces an enhancement to the self-supervised technique proposed in \cite{Liang2020SHOT}. CDCL augments this approach by incorporating intra-task information to generate task-relevant pseudo-labels. By leveraging such intra-task information, the model can better handle the complexities of continual learning in cross-domain settings, ensuring a more seamless and efficient adaptation to sequential tasks. CDCL's ability to handle both cross-domain and continual learning challenges makes it a promising solution for real-world applications requiring adaptive and lifelong learning capabilities.

\subsection{Continual learning}
\label{subsec:continual-learning}

Continual learning is the field where a single model, or ensemble of models, needs to sequentially learn a stream of tasks, whereby previous tasks become unavailable due to computational constraints, which impossibilities the re-fit of the whole dataset. Three scenarios \cite{vandeven2018generative} were proposed to standardize obtained results and enable fairer comparisons:

\begin{itemize}
    \item {\bf Task-incremental learning (TIL)}: The models receive a task identifier during inference. This is the most straightforward continual learning scenario, and typical solutions present a "multi-head" output layer to handle prediction.
    \item {\bf Domain-incremental learning (DIL)}: The task identifier is unavailable during inference. However, models do not need to infer the receiving task, as a single task is requested here, referred to as a domain. Typical solutions handle structures where the task is always the same, but the input distribution changes. This is the least explored scenario, as most researchers classify it as an intermediate stage between TIL and CIL.
    \item {\bf Class-incremental learning (CIL)}: The task identifier is available only during training time, and models must be able to solve and predict the receiving task during inference time. Typical solutions present a "single-head" output layer, as the model needs to learn to recognize new classes.
\end{itemize}

By defining clear evaluation protocols for TIL, DIL, and CIL, the current field can make significant progress towards developing more robust and adaptive continual learning algorithms for a wide range of real-world applications.

Furthermore, existing works are mainly divided into three categories:

\begin{itemize}
    \item {\bf Structure-based approach}: This approach tends to evolve the network by creating new nodes to handle incoming tasks while freezing the rest of the network parameters \cite{PNN, XDG}.

    \item {\bf Regularization-based approach}: These models penalize changes to parameters that are important to previously learned tasks by estimating their importance \cite{EWC, omniglot, SI}.
    
    \item {\bf Rehearsal-based approach}: This strategy replays old, or augmented, samples stored in memory when learning a new task \cite{GEM, A-GEM}. Strategies to optimize which samples to be stored in the memory are also typical \cite{GSS}. DER/DER++ \cite{DER} incorporates previous tasks {\it dark knowledge} to align past and current outputs. HAL \cite{HAL} adds additional replay objectives, reducing forgetting of key learned data points. Pseudo-rehearsal is an alternative, where an external generative model sequentially trains on all tasks and generates pseudo-samples from those data distributions \cite{DGR}.    
\end{itemize}

Current CL approaches are domain-specific and thus not adaptable to different but related unlabeled domains. The latter is a challenging problem, where a stream of tasks is generated from the labeled source domain and the label-free target domain. The proposed CDCL simultaneously addresses the task-incremental learning problem over multiple domains.

\label{subsec:proposed-method}

\section{Problem formulation}
\label{sec:problem-formulation}

Let $(\mathcal{D}_{\mathrm{S}_i}, \mathcal{D}_{\mathrm{T}_i})$ be an data stream system, with the data tuples $(\mathbf{x}_{\mathrm{S}_i}, \mathbf{y}_{\mathrm{S}_i})$ satisfying $(\mathbf{x}_{\mathrm{S}_i}, \mathbf{y}_{\mathrm{S}_i}) \stackrel{iid}{\sim} \mathcal{D}_{\mathrm{S}_i}(X_\mathrm{S}, Y_\mathrm{S})$, and $(\mathbf{x}_{\mathrm{T}_i})$ satisfying $(\mathbf{x}_{\mathrm{T}_i}) \stackrel{iid}{\sim} \mathcal{D}_{\mathrm{T}_i}(X_\mathrm{T})$. The data tuples contain a labeled set $X_{\mathrm{S}_i}$ paired with its target set $Y_{\mathrm{S}_i}$, along with an unlabeled input set $X_{\mathrm{T}_i}$, organized into sequential tasks $t_i \in \mathcal{T} = \{1,\cdots,T\}$, where the total number of tasks $T$ is unknown a priori. The goal is to learn a predictor $f: (\mathcal{X}_\mathrm{S} \cup \mathcal{X}_\mathrm{T}) \times \mathcal{T} \rightarrow \mathcal{Y}_\mathrm{T}$, which can be queried {\it at any time} to predict the target vector $y_\mathrm{T}$ associated to a sample in the target domain $x_\mathrm{T}$, where $(x_\mathrm{T}, y_\mathrm{T}) \sim P_t$. 

The solution to this problem is not trivial, as the model needs to handle {\it label scarcity in the target domain}, and also possibles {\it task drifts} $P_i(X_{\mathrm{S}}, Y_\mathrm{S}) \neq P_{i + 1}(X_\mathrm{S}, Y_\mathrm{S}) \wedge P_i(X_{\mathrm{T}}, Y_\mathrm{T}) \neq P_{i + 1}(X_\mathrm{T}, Y_\mathrm{T})$ and {\it domain drifts} $P_t(X_\mathrm{S}) \neq P_t(X_\mathrm{T}) \wedge P_t(Y_\mathrm{S}|X_\mathrm{S}) = P_t(Y_\mathrm{T}|X_\mathrm{T})$ along the training.

\section{Proposed method}
\label{sec:proposed-method}

\begin{figure}
    \centering
    \includegraphics{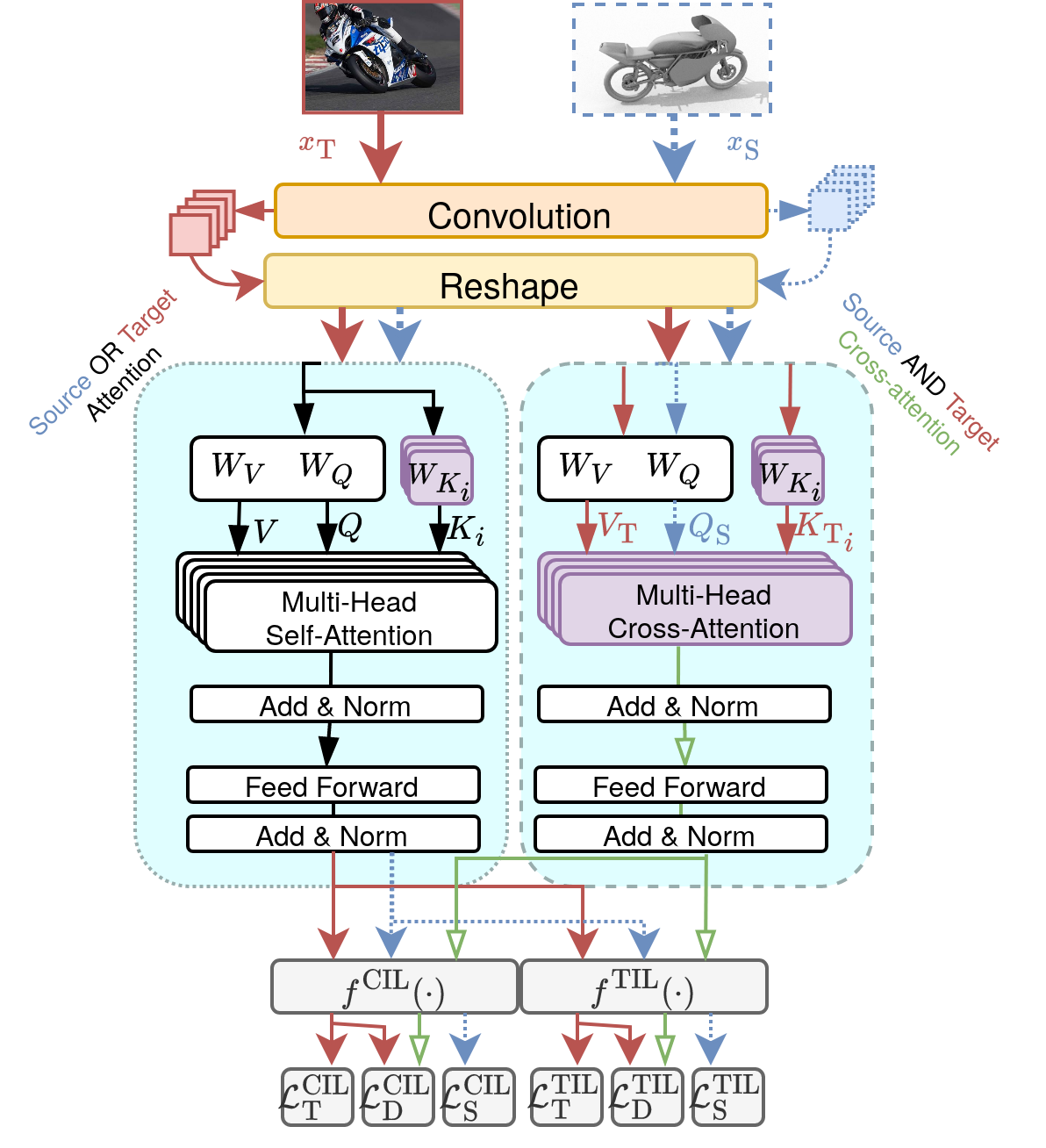}
    \caption{The proposed framework. The main contribution - the inter- intra-task cross attention - is highlighted with purple, which is responsible for aligning the source and target feature domains, and mitigating its feature-alignment catastrophic forgetting when new tasks arrive. When the network is faced with only a single input at a time, source (represented by the dashed blue line) or target (represented by the blocked red line), CDCL will process it via the self-attention mechanism (left side block). When the network is faced with both source and target inputs at the same time, CDCL will process them via the cross-attention mechanism (right side block), which outputs a mixed signal (represented by the blocked green open arrow). $\mathbf{b}_i$ is omitted for simplicity. The $f^\text{CIL}(\cdot)$ is a single-head output used for CIL scenarios along with the latest $\mathbf{K}_T$ and $\mathbf{b}_T$ instantiated. Meanwhile, the $f^\text{TIL}(\cdot)$ is a multi-head output used for TIL scenarios with the respective $\mathbf{K}_i$ and $\mathbf{b}_i$, as the task-identifier $t_i$ is provided.}
    \label{fig:cdcl}
\end{figure}

We propose a cross-domain task-incremental learning framework (CDCL) that alleviates the catastrophic forgetting of domain invariant features effectively. Figure \ref{fig:cdcl} presents an overview of the framework. The major contributions components follow:

\noindent {\bf 1}) {\it Inter- intra-task cross-attention} in the incremental transformer block implicitly retain previous tasks information into the attention maps.

\noindent {\bf 2}) {\it Intra-task center-aware pseudo-labeling} generates synthetic labels for the incoming unlabeled target inputs based on the intra-task information, integrating the same into the inter- intra-task cross-attention maps.

The rest of this section gives more details about the inter- intra-task cross-attention procedure; the intra-task center-aware pseudo-label strategy; the sample rehearsal configuration; the computational time complexity; and the upper-bound error of the proposed method.

\subsection{Inter- intra-task cross-attention mechanism}
\label{subsec:inter-task-attention-mechanism}

The proposed cross-attention mechanism is based on the CCT \cite{CCT} sequential pooling, an attention-based method that pools over the convolution tokenizer. Given an image $\mathbf{x}_i \in \mathbb{R}^{H \times W \times C}$ presented in task $t_i$:

\begin{equation}
    \label{eq:cnn}
    \mathbf{x}_{ct} = \textit{MaxPool}(\textit{ReLU}(\textit{Conv2d}(\mathbf{x}_i)))
\end{equation}

\noindent where $\textit{Conv2d}$ has $d$ filters, the same number as the embedding dimension of the transformer backbone - maintaining local spatial information. $\mathbf{x}_{ct}$ is the outputted convolution, replacing ViT's patch tokens. $\mathbf{x}_{ct}$ is then projected into global queries $\mathbf{Q}\in \mathbb{R}^{n \times d}$  and global values $\mathbf{V} \in \mathbb{R}^{n \times d}$. Furthermore, $\mathbf{x}_{ct}$ is also projected into a task-related keys $\mathbf{K}_i \in \mathbb{R}^{n \times d}$ and task-related bias $\mathbf{b}_i \in \mathbb{R}^{1 \times n}$. 

The self-attention operation is now applied:

\begin{equation}
    \mathbf{x}_{Li} = \frac{\mathbf{Q}\mathbf{K}_i^T + \mathbf{b}_i}{\sqrt{d}}\mathbf{V} \in \mathbb{R}^{b \times n \times d}
\end{equation}

\noindent where $\mathbf{x}_{Li}$ is the output of an $L$ layer transformer encoder, $b$ is the batch size, $n$ is the senquence length, and $d$ is the total embedding dimension. The cross-attention module is derived from the self-attention module. The difference is that the input of the cross-attention is a pair of images, \textit{i.e.} $\mathbf{x}_{\mathrm{S}_i}$ and $\mathbf{x}_{\mathrm{T}_i}$. The cross-attention module can be calculated as follows:

\begin{equation}
    \label{eq:attention}
    \mathbf{x}_{Li} = \frac{\mathbf{Q}_\mathrm{S}\mathbf{K}_i^T + \mathbf{b}_i}{\sqrt{d}}\mathbf{V}_\mathrm{T} \in \mathbb{R}^{b \times n \times d}
\end{equation}

\noindent where $\mathbf{Q}_\mathrm{S}$ are the global queries from image $\mathbf{x}_{\mathrm{S}_i}$ and $\mathbf{V}_\mathrm{T}$ are the global values from image $\mathbf{x}_{\mathrm{T}_i}$. Note that $\mathbf{x}_{\mathrm{T}_i}$ convolution token is also projected into the task-related $\mathbf{K}_i$ and $\mathbf{b}_i$, producing an inter-task projection. A new pair $\mathbf{K}_i$ and $\mathbf{b}_i$ is created when a new task arrives, while previously learned $\mathbf{K}_{1,\cdots,i}$ and $\mathbf{b}_{1,\cdots,i}$ are frozen, preserving the knowledge from the previous feature aligned tasks and building the current feature aligned knowledge based on the global $\mathbf{Q}$ and $\mathbf{V}$.

The attention maps are used to generate an importance weighting for each input convolution token:

\begin{equation}
    \mathbf{x}_{Li}^\prime = \text{softmax} (g(\mathbf{x}_{Li})^T) \in \mathbb{R}^{b \times 1 \times n}
\end{equation}

\begin{equation}
    \mathbf{z}_i = \mathbf{x}_{Li}^\prime \mathbf{x}_{Li} = \text{softmax}(g(\mathbf{x}_{Li})^{T}) \times \mathbf{x}_{Li}
\end{equation}

Throughout the paper, this whole operation is simplified by the given notation:

\begin{equation}
    a(\mathbf{x}_i) = \mathbf{z}_i \in \mathbb{R}^{b \times 1 \times d}
\end{equation}

The final feature output is generated by flattening $\mathbf{z}_i \in \mathbb{R}^{b \times d}$. These features can be sent through a classifier for both training and evaluation. CDCL evaluates $\mathbf{z}_i$ in both inter-task (CIL) and intra-task (TIL) manner:

\begin{equation}
    \mathbf{y}^\text{TIL} = f^\text{TIL}(z_i) \in \mathbb{R}^{b \times u}
\end{equation}

\begin{equation}
    \mathbf{y}^\text{CIL} = f^\text{CIL}(z_i) \in \mathbb{R}^{b \times u}
\end{equation}

\noindent where $u$ is the number of classes in the current task. From here, the following objective functions are applied:

\begin{equation}
    \label{eq:loss_s_cil}
    \mathcal{L}_\mathrm{S}^\text{CIL} = - \sum \mathbf{y_\mathrm{S}} \log(\mathbf{y}_\mathrm{S}^\text{CIL}) \in \mathbb{R}^{b}
\end{equation}

\begin{equation}
    \label{eq:loss_t_cil}
    \mathcal{L}_\mathrm{T}^\text{CIL} = - \sum \mathbf{y_\mathrm{S}} \log(\mathbf{y}_\mathrm{T}^\text{CIL}) \in \mathbb{R}^{b}
\end{equation}

\begin{algorithm}[t]
\caption{CDCL learning pseudo-algorithm}
\label{alg:algorithm}
\KwIn{$\mathcal{D}_\mathrm{S}, \mathcal{D}_\mathrm{T}$}
\KwOut{$f: (\mathbf{x}_\mathrm{S} \cup \mathbf{x}_\mathrm{T}) \times \mathcal{T} \rightarrow \mathbf{y}_\mathrm{T}$}

\For{$t_i$ in $\mathcal{T}$} {
    $\mathbf{x}_\mathrm{S}, \mathbf{y}_\mathrm{S} \leftarrow \mathcal{D}_{\mathrm{S}_i}$; \\
    $\mathbf{x}_\mathrm{T} \leftarrow \mathcal{D}_{\mathrm{T}_i}$; \\
    Random initialize $K_i$ and $b_i$; \\
    \For{epoch in epochs}{
        $\mathcal{L}^\text{CIL} \leftarrow \mathcal{L}^\text{TIL} \leftarrow \mathcal{L}_\mathrm{R} \leftarrow 0$;\\
        \If{epoch $<$ warm-up-epochs}{
            $\mathcal{L}_\mathcal{S}^\text{CIL} \leftarrow - \sum \mathbf{y_\mathrm{S}} \log(\mathbf{y}_\mathrm{S}^\text{CIL})$; \# Eq.\ref{eq:loss_s_cil} \\
            $\mathcal{L}_\mathcal{S}^\text{TIL} \leftarrow - \sum \mathbf{y_\mathrm{S}} \log(\mathbf{y}_\mathrm{S}^\text{TIL})$; \# Eq.\ref{eq:loss_s_til} \\
        }
        \Else{
            Compute $\mathbf{c}_k$ and $\hat{\mathbf{y}}_\mathrm{T}$; \# Eqs. \ref{eq:initcenter} and \ref{eq:initlabel} \\
            Compute $\mathbb{P}$; \# Eq. \ref{eq:p} \\
                $\mathcal{L}^\text{CIL} \leftarrow \mathcal{L}_\mathrm{S}^\text{CIL} + \mathcal{L}_\mathrm{T}^\text{CIL} + \mathcal{L}_\mathrm{D}^\text{CIL}$; \# Eq.\ref{eq:loss_cil} \\
                $\mathcal{L}^\text{TIL} \leftarrow \mathcal{L}_\mathrm{S}^\text{TIL} + \mathcal{L}_\mathrm{T}^\text{TIL} + \mathcal{L}_\mathrm{D}^\text{TIL}$; \# Eq.\ref{eq:loss_til} \\
                \If{$t_i > 1$}{
                    {$\mathcal{L}_R \leftarrow \mathcal{L}_R^\mathrm{ST} + \mathcal{L}_R^\mathrm{D} + \mathcal{L}_R^\mathrm{Z}$; \# Eq.\ref{eq:loss_r}} \\
                }
        }
        $\mathcal{L} \leftarrow \mathcal{L}^\text{CIL} + \mathcal{L}^\text{TIL} + \mathcal{L}_\mathrm{R}$; \\
        $\theta \leftarrow \theta - \lambda \nabla_\theta \mathcal{L}$; \# Update model parameters
    }
    Stores $\lfloor |\mathcal{M}|/t_i \rfloor$ records per task in memory $\mathcal{M}$; \\
}
\end{algorithm}

\begin{equation}
    \label{eq:loss_d_cil}
    \mathcal{L}_\mathrm{D}^\text{CIL} = \sum \mathbf{y}_\mathrm{S+T}^\text{CIL} \log(\mathbf{y}_\mathrm{T}^\text{CIL}) \in \mathbb{R}^{b}
\end{equation}

\noindent where $\mathbf{y_\mathrm{S}}$ is the label vector from the source domain, $\mathbf{y}_\mathrm{S}^\text{CIL}, \mathbf{y}_\mathrm{T}^\text{CIL}$ are respectively the output vector of the source input and target input through the intra-task classifier, and $\mathbf{y}_\mathrm{S+T}^\text{CIL}$ is the output vector of the inter- intra-task cross-attention applied to both the source and target image vector through the intra-task classifier, i.e., they map the source and target classes information between the tasks. Similar objectives are optimized through the inter-task classifier, which maps the source and target classes information within a task:

\begin{equation}
    \label{eq:loss_s_til}
    \mathcal{L}_\mathrm{S}^\text{TIL} = - \sum \mathbf{y_\mathrm{S}} \log(\mathbf{y}_\mathrm{S}^\text{TIL}) \in \mathbb{R}^{b}
\end{equation}

\begin{equation}
    \label{eq:loss_t_til}
    \mathcal{L}_\mathrm{T}^\text{TIL} = - \sum \mathbf{y_\mathrm{S}} \log(\mathbf{y}_\mathrm{T}^\text{TIL}) \in \mathbb{R}^{b}
\end{equation}

\begin{equation}
    \label{eq:loss_d_til}
    \mathcal{L}_\mathrm{D}^\text{TIL} = \sum \mathbf{y}_\mathrm{S+T}^\text{TIL} \log(\mathbf{y}_\mathrm{T}^\text{TIL}) \in \mathbb{R}^{b}
\end{equation}

We will refer to the inter-task and intra-task losses as single objectives throughout the paper:

\begin{equation}
    \label{eq:loss_cil}
    \mathcal{L}^\text{CIL} = \mathcal{L}_\mathrm{S}^\text{CIL} + \mathcal{L}_\mathrm{T}^\text{CIL} + \mathcal{L}_\mathrm{D}^\text{CIL} \in \mathbb{R}^{b}
\end{equation}

\begin{equation}
    \label{eq:loss_til}
    \mathcal{L}^\text{TIL} = \mathcal{L}_\mathrm{S}^\text{TIL} + \mathcal{L}_\mathrm{T}^\text{TIL} + \mathcal{L}_\mathrm{D}^\text{TIL} \in \mathbb{R}^{b}
\end{equation}

\subsection{Intra-task center-aware pseudo-label}
\label{subsec:intra-task-center-aware-pseudo-label}

To ensure a feature alignment on correct samples, we build a set $\mathbb{P}$ containing pairs of similar inputs from the source and target domain. $\mathbb{P}$ is built with both feature similarity and label similarity in mind. To improve inter-task alignment, we increment \cite{Liang2020SHOT} by adding intra-task knowledge into their center-aware pseudo-label mechanism, i.e., we use only information extracted from the current task to build the current task centroids. When a new task arrives, a warm-up stage starts where the classifiers will train solely in the source domain inputs. After the warm-up stage finishes, we build category centroids in the target domain by weighted k-means clustering based on the intra-task classification information, updating it at every training epoch\footnote{We recreate the centroids at every training epoch.}:

\begin{equation}
    \label{eq:initcenter}
    \mathbf{c}_k = \frac{\sum_{\mathcal{D}_{\mathrm{T}_i}} \mathbf{y}_\mathrm{T}^{\text{TIL}} a(\mathbf{x}_\mathrm{T})}{\sum_{\mathcal{D}_{\mathrm{T}_i}}\mathbf{y}_\mathrm{T}^{\text{TIL}}}
\end{equation}

\noindent where $\mathbf{y}_\mathrm{T}^\text{TIL}$ corresponds to the intra-task prediction on the target input samples and $k$ refers to the number of classes in the current task. Pseudo-labels of the target data are produced via the nearest neighbor classifier:

\begin{equation}
    \label{eq:initlabel}
    \hat{\mathbf{y}}_\mathrm{T} = \arg\min_k d(\mathbf{c}_k, a(\mathbf{x}_\mathrm{T}))
\end{equation}

\noindent where $d(\cdot, \cdot)$ is a distance metric, as cosine similarity or Euclidean distance. Finally, the set $\mathbb{P}$ is updated accordingly to the pseudo-labels, discarding noise and only allowing paired records that match the source input label with the target-generated pseudo-label:

\begin{equation}\small
    \label{eq:p}
    \begin{aligned}
        \mathbb{P} = \{(x_S,y_S,x_T)| & \forall x_S \in a(X_S), \\ & \forall x_T \in a(X_T), \\ & \arg \min_{(x_S, y_S, x_T)} d(x_S, x_T) \wedge y_S = \hat{y}_T\}
    \end{aligned}
\end{equation}



\subsection{Sample rehearsal}
\label{subsec:sample-rehearsal}

Although neural networks are susceptible to catastrophic forgetting, it is believed that some residual from past experiences can still be found within its weights. Therefore, rehearsing samples is an effective strategy to drive these residuals toward previous knowledge presentation. However, memory size is constrained by having a fixed size or growing very slowly with the number of incoming tasks. Therefore, it is important to select samples that better describe previous feature alignment knowledge, hence having a better regeneration over residuals. At the end of the training of the current task $t_i$, $\lfloor |\mathcal{M}|/t_i \rfloor$ records\footnote{Each record consist of the tuple ($\mathbf{x}_\mathrm{S}, \mathbf{x}_\mathrm{T}, \mathbf{y}_\mathrm{S}, \mathbf{y}_\mathrm{S}^\text{CIL}, \mathbf{y}_\mathrm{T}^\text{CIL})$ in memory with respective notation $(\mathbf{x}_\mathrm{S}^\mathrm{R}, \mathbf{x}_\mathrm{T}^\mathrm{R}, \mathbf{y}^\mathrm{R}, \mathbf{y}_\mathrm{S}^\mathrm{R}, \mathbf{y}_\mathrm{T}^\mathrm{R})$ where $\mathrm{R}$ denotes "Rehearsal".} with highest intra-task confidence, $max(\mathbf{y}_S^{TIL}) \vee \max(\mathbf{y}_T^{TIL})$, are stored in the memory $\mathcal{M}$.

To preserve the knowledge about previous tasks, we seek to minimize the following losses:

\begin{equation}
    \mathcal{L}_R^\mathrm{ST} = - \sum \mathbf{y^\mathrm{R}} \log(f^\text{CIL}(a(\mathbf{x}_\mathrm{S}^\mathrm{R})) f^\text{CIL}(a(\mathbf{x}_\mathrm{T}^\mathrm{R}))) \in \mathbb{R}^{b}
\end{equation}


\begin{equation}
    \mathcal{L}_R^D = \sum f^\text{CIL}(a(\mathbf{x}_\mathrm{S}^\mathrm{R}), a(\mathbf{x}_\mathrm{T}^\mathrm{R})) f^\text{CIL}(a(\mathbf{x}_\mathrm{T}^\mathrm{R})) \in \mathbb{R}^{b}
\end{equation}

\noindent where $\mathbf{x}_\mathrm{S}^\mathrm{R}, \mathbf{x}_\mathrm{T}^\mathrm{R}, \mathbf{y}^\mathrm{R}$ are respectively the source image, target image and original source labels recorded in the memory $\mathcal{M}$. Additionally, CDCL also attempts to maintain previously learned task boundaries by logit replay:

\begin{equation}
    \mathcal{L}_R^\mathrm{Z} = \sum \mathbf{y}_\mathrm{S}^\mathrm{R} \log( \frac{\mathbf{y}_\mathrm{T}^\mathrm{R}}{f^\text{CIL}(a(\mathbf{x}_\mathrm{T}^\mathrm{R}))}\frac{\mathbf{y}_\mathrm{S}^\mathrm{R}}{f^\text{CIL}(a(\mathbf{x}_\mathrm{S}^\mathrm{R}))}) \in \mathbb{R}^{b}
\end{equation}


\noindent where $\mathbf{y}_\mathrm{S}^\mathrm{R}, \mathbf{y}_\mathrm{T}^\mathrm{R}$ are respectively the source logits and target logits recorded in the memory $\mathcal{M}$.

Finally, we consolidated the rehearsal objective\footnote{Note that the pseudo-label centroid were built using intra-task knowledge (information within a task), but the rehearsal loss apply it via inter-task outputs (information between all tasks)} as:

\begin{equation}
    \label{eq:loss_r}
    \mathcal{L}_R = \mathcal{L}_R^\mathrm{ST} + \mathcal{L}_R^\mathrm{D} + \mathcal{L}_R^\mathrm{Z} \in \mathbb{R}^{b}
\end{equation}

Algorithm \ref{alg:algorithm} depicts a simplified representation of how and when CDCL optimizes each of its objectives.

\subsection{Time Complexity Analysis}

The computational time complexity of the proposed method is derived into two main procedures: the convolution tokenizer operation; and the inter- intra-task cross-attention operation.

The convolution tokenizer in equation \ref{eq:cnn} performs $O(n)$ operations per layer $L_c$, where $n$ is the input feature space size of $x_i$. A cross-attention operation has a linear transformation of its input into the three projections $\mathbf{Q}, \mathbf{V},$ and $\mathbf{K}$, followed by a post-multiplication of shape $(d \times d)$, resulting in $O(nd^2)$. Computing the output of the attention mechanism following equation \ref{eq:attention} has complexity $O(dn^2)$ to compute $\frac{\mathbf{Q}\mathbf{K}_i^T + \mathbf{b}_i}{\sqrt{d}}$, and $O(nd^2)$ for the post-multiplication with $\mathbf{V}$, repeated by $L_a$ layers. Therefore, a single forward-pass time complexity of CDCL is defined by:

\begin{equation}
    O(\underbrace{nL_c}_{\text{Conv Token}} + \underbrace{(dn^2 + nd^2)L_a}_{\text{Cross-attentions}})
\end{equation}

This time complexity equation doesn't take into consideration the impact of general-purpose hardware architectures in the algorithm, which explains how transformers (and consequently CDCL) run faster than previous
networks, even though it presents a higher computational complexity \cite{Transformers}.

\subsection{Theoretical analysis}
\label{subsec:theoretical-analysis}

In this section, we describe the error bound for the general cross-domain continual learning problem and how it influenced the proposed method architecture:

\begin{theorem}
\label{theorem_1}
{\it Theorem 1}: Given two domain distributions $\mathcal{D}_\mathrm{S}(X_\mathrm{S})$ and $\mathcal{D}_\mathrm{T}(X_\mathrm{T})$, the target domain error $\varepsilon_\mathrm{T}$ is bound by \cite{ben2010theory}:

\begin{equation}\small
    d_{\mathcal{H}\Delta\mathcal{H}}(X_\mathrm{S}, X_\mathrm{T}) = \underset{\eta \in \mathcal{H}}{2sup}\left|{P[\eta(X_\mathrm{S})\!=\!1]} - {P[\eta(X_\mathrm{T})\!=\!1]}  \right|
\end{equation}

\begin{equation}
\label{eq:ben2010theorem-1}
    \varepsilon_\mathrm{T} \leq \varepsilon_\mathrm{S} + d_{\mathcal{H}\Delta\mathcal{H}}(X_\mathrm{S}, X_\mathrm{T}) + C^*
\end{equation}

\noindent where $d_{\mathcal{H}\Delta\mathcal{H}}(X_\mathrm{S}, X_\mathrm{T})$ is the $\mathcal{H}\Delta\mathcal{H}$ divergence, which relies on the capacity of the hypothesis class $\mathcal{H}$ to distinguish between examples generated by $\mathcal{D}_\mathrm{S}(X_\mathrm{S})$ from examples generated by $\mathcal{D}_\mathrm{T}(X_\mathrm{T})$. $C^*$ is the error of an optimal classifier for both source domain $\mathcal{D}_\mathrm{S}$ and target domain $\mathcal{D}_\mathrm{T}$.

\end{theorem}

\begin{theorem}
\label{theorem_2}
{\it Theorem 2}: For every new sequential task $t_i \in \mathcal{T} = \{1,\cdots,T\}$ and incoming domain distributions $\mathcal{D}_{\mathrm{S}_i}(X_\mathrm{S})$ and $\mathcal{D}_{\mathrm{T}_i}(X_\mathrm{T})$, let $\lambda_i = d_{\mathcal{H}\Delta\mathcal{H}}(\mathbf{z}_{\mathrm{S}_i}, \mathbf{z}_{\mathrm{T}_i})$ be the domain discrepancy of the feature distributions $\mathbf{z}_{\mathrm{S}_i}$ and $\mathbf{z}_{\mathrm{T}_i}$. The target domain error $\varepsilon_{\mathrm{T}_i}$ at task $t_i$ is bound by:

\begin{equation}
    \varepsilon_{\mathrm{T}_i} \leq \varepsilon_{\mathrm{S}_i} + \lambda_i + C_i^*
\end{equation}

\noindent where $C_i^* = \min_\theta(\varepsilon_{\mathrm{S}_i} + \varepsilon_{\mathrm{T}_i})$ is the error of an optimal classifier for both source domain $\mathcal{D}_{\mathrm{S}_i}$ and target domain $\mathcal{T}_{\mathrm{S}_i}$ at task $t_i$.

\end{theorem}

\begin{theorem}
\label{theorem_3}
{\it Theorem 3}: For every new sequential task $t \in \mathcal{T} = \{1,\cdots,T\}$ and incoming domain distributions $\mathcal{D}_{\mathrm{S}_i}(X_\mathrm{S})$ and $\mathcal{D}_{\mathrm{T}_i}(X_\mathrm{T})$, let $\lambda_i = d_{\mathcal{H}\Delta\mathcal{H}}(\mathbf{z}_{\mathrm{S}_i}, \mathbf{z}_{\mathrm{T}_i})$ be the domain discrepancy of the feature distributions $\mathbf{z}_{\mathrm{S}_i}$ and $\mathbf{z}_{\mathrm{T}_i}$, and $P_{\mathrm{R}_i}, P_{\mathcal{M}_i}$ denote respectively the conditional distributions of labels $\mathbf{y}_{\mathrm{S}_i}$ given the raw features $\mathbf{z}_{\mathrm{S}_i}$ and memory stored features $\mathbf{z}_{\mathcal{M}_i}$. The total error of the target domain error $\varepsilon_{\mathrm{T}}$ at task $t$ is bound by:

\begin{equation}
    \varepsilon_{\mathrm{T}} \leq \sum_{i=1}^{t}(\varepsilon_{\mathrm{S}_i} + \lambda_i) + \sum_{i=1}^{t-1} \text{KL}(P_{\mathcal{M}_i} || P_{\mathrm{R}_i}) + C^*
\end{equation}

\noindent where $C^* = \min_\theta \sum_{i=1}^{t}(\varepsilon_{\mathrm{S}_i} + \varepsilon_{\mathrm{T}_i})$ is the error of an optimal classifier for both source domain $\mathcal{D}_{\mathrm{S}_\mathcal{T}}$ and target domain $\mathcal{D}_{\mathrm{T}_\mathcal{T}}$ across all tasks. Here, we choose the KL-divergence due to its simplification and similarity to other common information theory equations, but any divergence equation following Eq. \ref{eq:ben2010theorem-1} can be used.
    
\end{theorem}

Theorem 2 assists us to prove theorem 3, which is proved by \cite{TwoStreamCL} for a similar problem and adapted here.

\begin{proof}
\label{proof_1}
{\it Proof 1}: At task $t$, the error of previous learned target experiences $\mathcal{D}_{\mathrm{T}_i}(X_\mathrm{T})$ for $i \in \{1,\cdots,t - 1\}$ is estimated by $\hat{\varepsilon_{\mathrm{T}_i}}$ since we rehearse $\mathbf{z}_{\mathcal{M}_i}$ to mimic $\mathbf{z}_{\mathcal{S}_i}$ during training. The total error of the target domain error $\varepsilon_{\mathrm{T}}$ at task $t_i$ is:

\begin{equation}
    \begin{aligned}
        \varepsilon_\mathrm{T} & = \varepsilon_{\mathrm{T}_t} + \sum_{i=1}^{t-1} \hat{\varepsilon_{\mathrm{T}_i}} \\
        & \leq \varepsilon_{\mathrm{S}_t} + \lambda_t + \sum_{i=1}^{t-1} (\hat{\varepsilon_{\mathrm{S}_i}} + \lambda_i) + C^* \cdot
    \end{aligned}
\end{equation}

Additionally, the divergence between $P_{\mathrm{R}_i}$ and $P_{\mathcal{M}_i}$ can be understood as:

\begin{equation}
    \begin{aligned}
        \text{KL} &
        \big(P_{\mathcal{M}_i} (\mathbf{y}_{\mathrm{S}_i}|\mathbf{z}_{\mathcal{M}_i}) || P_{\mathrm{R}_i}(\mathbf{y}_{\mathrm{S}_i}|\mathbf{z}_{\mathrm{S}_i})\big) \\ 
        & = \sum_{y \in \mathbf{y}} P_{\mathcal{M}_i}(\mathbf{y}_{\mathrm{S}_i}|\mathbf{z}_{\mathcal{M}_i}) \log \frac{P_{\mathcal{M}_i}(\mathbf{y}_{\mathrm{S}_i}|\mathbf{z}_{\mathcal{M}_i})}{P_{\mathrm{R}_i}(\mathbf{y}_{\mathrm{S}_i}|\mathbf{z}_{\mathrm{S}_i})} \\
        & = \mathbb{E} \Big[ \log \frac{P_{\mathcal{M}_i}(\mathbf{y}_{\mathrm{S}_i}|\mathbf{z}_{\mathcal{M}_i})}{P_{\mathrm{R}_i}(\mathbf{y}_{\mathrm{S}_i}|\mathbf{z}_{\mathrm{S}_i})} \Big] \\
        & = \mathbb{E} [\log{P_{\mathcal{M}_i}(\mathbf{y}_{\mathrm{S}_i}|\mathbf{z}_{\mathcal{M}_i})}] - \mathbb{E} [\log{P_{\mathrm{R}_i}(\mathbf{y}_{\mathrm{S}_i}|\mathbf{z}_{\mathrm{S}_i})}] \\
        & = \hat{\varepsilon_{\mathrm{S}_i}} - \varepsilon_{\mathrm{S}_i} \cdot
    \end{aligned}    
\end{equation}

Hence, we can estimate the error for each source domain $\varepsilon_{\mathrm{S}_i}$ in a specific task $t$ as:

\begin{equation}
    \varepsilon_{\mathrm{S}_t} = \hat{\varepsilon_{\mathrm{S}_t}} - \text{KL}\big(P_{\mathcal{M}_t} || P_{\mathrm{R}_t}\big)
\end{equation}

Finally, we can estimate the total error bound for the target domain $\varepsilon_{\mathrm{T}}$ at a task $t$ by:

\begin{equation}\small
    \begin{aligned}
        \varepsilon_{\mathrm{T}} & \leq \varepsilon_{\mathrm{S}_t} + \lambda_t + \sum_{i=1}^{t-1}(\hat{\varepsilon_{\mathrm{S}_i}} + \lambda_i) +  C^* \\
        & = \varepsilon_{\mathrm{S}_t} + \lambda_t + \sum_{i=1}^{t-1}(\varepsilon_{\mathrm{S}_i} + \text{KL}\big(P_{\mathcal{M}_i} || P_{\mathrm{R}_i}\big) + \lambda_i) +  C^* \\
        & = \sum_{i=1}^t(\varepsilon_{\mathrm{S}_i} + \lambda_i) + \sum_{i=1}^{t-1} \text{KL}\big(P_{\mathcal{M}_i} || P_{\mathrm{R}_i}\big) + C^* \cdot
    \end{aligned}
\end{equation}

\end{proof}

Finally, we can use theorem 3 to explain the objective choices in CDCL based on proof 1: $\mathcal{L}_\mathrm{T}^\text{CIL}, \mathcal{L}_\mathrm{D}^\text{CIL}, \mathcal{L}_\mathrm{T}^\text{TIL}, \mathcal{L}_\mathrm{D}^\text{TIL}$ 
contribute to the creation of a feature-aligned domain-invariant latent space, being related to $\lambda_i$; $\mathcal{L}_\mathrm{R}$ is related to the $\text{KL}$ term, which is minimized by the samples chosen by the intra-task center-aware pseudo-label approach; and finally, $\mathcal{L}_\mathrm{S}^\text{CIL}, \mathcal{L}_\mathrm{S}^\text{TIL}$ fits CDCL in the source domain, decreasing the incremental source domain error rate $\varepsilon_{\mathrm{S}_t}$. $C^*$ represents how suitable the model architecture is to find an optimal solution for both data domains.





\section{Experiments}
\label{sec:experiments}

\subsection{Benchmarks}
\label{subsec:benchmarks}

The proposed method is verified in five popular UDA benchmarks, including VisDA-2017 \cite{VisDA-2017}, Office-Home \cite{Office-Home}, Office-31 \cite{Office-31}, DomainNet \cite{DomainNet}, and MNIST$\leftrightarrow$USPS \cite{MNIST, USPS}.

\textbf{MNIST$\leftrightarrow$USPS}: Gray-scale images of hand-written digits. The 10 classes are split into 5 tasks of 2 classes each.

\textbf{VisDA-2017}: Consists of a source dataset of synthetic 2D renderings of 3D models generated from different angles and with different lighting conditions and a target dataset of photo-realistic images. The 12 classes are split into 4 tasks of 3 classes each.

\textbf{DomainNet}: Contains 345 classes of ordinary objects, represented in six domains: Clipart (clp), Infographics (inf), Painting (pnt), Quickdraw (qdr), Real (rel), and Sketch (skt). The 345 classes are divided into 15 tasks of 23 classes each.

\textbf{Office31}: Contains 31 object categories in three domains: Amazon (A), DSLR (D), and Webcam (W). The 31 categories in the dataset consist of objects commonly encountered in office settings. To ease the baseline setup, the category "trash can" was dropped, making this a problem of 30 classes divided into 5 tasks of 6 classes each.

\textbf{Office-Home}: Consists of 65 categories on four domains: Art (Ar), Clipart (Cl), Product (Pr), and Real-World (Re). The 65 classes are divided into 13 tasks of 5 classes each.

\subsection{Baselines}
\label{subsec:baselines}

We compare our method with the state-of-the-art method on UDA tasks, CDTrans-S and CDTrans-B \cite{CDTrans}, as well as with state-of-the-art methods in continual learning, such as DER \cite{DER}, DER++ \cite{DER}, and HAL \cite{HAL}. We also compare our method against MLS \cite{Simon_2022_CVPR}, which perform supervised cross-domain continual learning. Finally, we showcase the performance of the Transferable Vision Transformer (TVT) \cite{TVT}, another state-of-the-art UDA method, as a full trainig upper bound baseline, clearly depicting the catastrophic forgetting gap between static learning and continual learning.

We created two instances of CDCL for the experiments: A small-sized instance for the MNIST$\leftrightarrow$USPS experiments and a larger one for all the other benchmarks. The small version of CDCL had 7 transformer encoder layers, a 2-layer convolution tokenizer with a 7x7 kernel size, and a convolution input feature space of size 28x28x1. The larger version had 14 transformer encoder layers, a 2-layer convolution tokenizer with a 7x7 kernel size, and a convolution input feature space of size 224x224x3. 

All baselines were executed over 125 epochs (25 warm-up epochs and 25 cooldown epochs) with a maximum and fixed memory size of 1000 records. CDTrans-S and CDTrans-B were performed with the same hyper-parameters described in the original paper \cite{CDTrans}. DER, DER++, and HAL were performed with the best hyper-parameters described in the Mammoth library \cite{boschini2022class}. CDCL uses AdamW optimizer \cite{AdamW} with a warm-up learning-rate $\lambda = 10^{-5}$, a cosine annealing learning-rate starting at $\lambda = 5* 10^{-5}$ and a minimum learning-rate of $\lambda = 10^{-6}$. All experiments were executed in an NVIDIA GeForce RTX 2080Ti with 11GB of VRAM.

\subsection{Metrics}
\label{subsec:metrics}

The continual learning protocol is followed, where we observe two metrics: 

\begin{equation}\small
    \label{eq:ACC}
    \textbf{\text{Average Accuracy: }} \text{ACC}_\uparrow = \frac{1}{T} \sum_{i=1}^{T} R_{T,i}
\end{equation}



\begin{equation}\small
    \label{eq:FGT}
    \textbf{\text{Forgetting: }} \text{FGT}_\downarrow = \frac{1}{T - 1} \sum_{i=1}^{T - 1} \max\limits_{i \in {1,\cdots,T-1}}(R_{i,i}  - R_{T,i})
\end{equation}

\noindent where $R \in \mathbb{R}^{T \times T}$ is a test classification matrix on the target domain, with $R_{i,j}$ representing the accuracy in task $t_j$ after thoroughly learning $t_i$. The details are given by \cite{GEM, Chaudhry2018}.

\subsection{Numerical results}
\label{subsec:numerical-results}

\begin{table*}[t]
\setlength\tabcolsep{1.0pt}
    \begin{tabularx}{\linewidth}{Y|Y|YYYYYYYYYY}
    \hline
    \multicolumn{2}{c|}{Method} &
    A$\to$D & 
    A$\to$W & 
    D$\to$A & 
    D$\to$W & 
    W$\to$A & 
    W$\to$D & 
    MN$\to$US & 
    US$\to$MN & 
    VisDA-2017 \\ 
    
    \hline 
    
    DER      & \multirow{8}{*}{\rotatebox[origin=c]{90}{\textbf{TIL}}} & 4.45 & 4.20 & 3.99 & 11.50 & 4.29 & 9.14 & 62.79 & 73.16 & \underline{11.43} \\
    DER++    & & 4.21 & 4.18 & 4.21 & 8.51 & 3.74 & 8.42 & 79.81 & 73.24 & 10.24 \\
    HAL      & & 4.62 & 5.02 & 3.93 & 5.77 & 3.70 & 5.28 & \underline{80.97} & \underline{73.38} & 9.84 \\
    MSL      & & 4.75 & 4.10 & 3.89 & \underline{11.65} & 4.67 & \underline{9.23} & 63.01 & 71.43 & 12.21 \\
    \cline{0-0} \cline{3-11}
    CDTrans-S & & 2.50 & \underline{5.20} & 4.50 & 5.20 & 4.50 & 2.50 & 10.78 & 10.68 & 8.90 \\
    CDTrans-B & & \underline{5.96} & 3.90 & \underline{4.70} & 3.60 & \underline{4.70} & 5.80 & 8.77 & 9.42 & 7.85 \\
    \cline{0-0} \cline{3-11}
    Ours (ACC) & & \textbf{26.22} & \textbf{22.43} & \textbf{28.74} & \textbf{55.44} & \textbf{26.54} & \textbf{53.21} & \textbf{91.91} & \textbf{81.48} & \textbf{40.80} \\
    Ours (FGT) & & 5.95 & 6.86 & 0.33 & 6.38 & 3.41 & 9.38 & 7.38 & 10.22 & 7.94 \\
    
    \hline
    \hline
    
    DER      & \multirow{6}{*}{\rotatebox[origin=c]{90}{\textbf{CIL}}} & 2.07 & 3.62 & 5.23 & 24.81 & \underline{7.34} & \underline{26.50} & 29.71 & 29.34 & 10.09 \\
    DER++    & & 2.07 & 3.75 & \underline{6.36} & 11.11 & 4.43 & 19.25 & \underline{40.82} & \underline{33.06} & 9.32 \\
    HAL      & & \underline{4.35} & \underline{5.81} & 3.23 & 5.43 & 4.43 & 11.18 & 29.08 & 25.94 & 8.50 \\
    MSL      & & 3.23 & 4.42 & 4.43 & \underline{25.23} & \textbf{8.01} & 24.20 & 30.22 & 30.56 & \textbf{12.23} \\
    \cline{0-0} \cline{3-11}
    Ours (ACC) & & \textbf{9.68} & \textbf{10.98} & \textbf{7.02} & \textbf{27.97} & 6.73 & \textbf{28.98} & \textbf{66.73} & \textbf{52.50} & \underline{10.16} \\
    Ours (FGT) & & 4.14 & 4.37 & 1.19 & 8.30 & 1.54 & 7.20 & 37.03 & 27.81 & 26.20 \\
    
    \hline
    \hline

    \multicolumn{2}{c|}{TVT (Static UDA)} & 96.39 & 96.35 & 84.91 & 99.37 & 86.05 & 100.0 & 98.26 & 99.70 & 83.92 \\

    \bottomrule
    
    \end{tabularx}
\vspace{1mm}
\caption{Comparison with SoTA methods' ACC on Office-31, MNIST$\leftrightarrow$USPS, and VisDA-2017. The best result in each experiment is highlighted with bold font, while the second-best is highlighted with underline font.}
\vspace{-3mm}
\label{tab:office31}
\end{table*}

\begin{table*}[h]
\setlength\tabcolsep{1.0pt}
    \begin{tabularx}{\linewidth}{YY|Y|YYYYYYYYYYYY}
    \hline
    \multicolumn{3}{c|}{Method} &
      Ar$\to $Cl &
      Ar$\to $Pr &
      Ar$\to $Re &
      Cl$\to $Ar &
      Cl$\to $Pr &
      Cl$\to $Re &
      Pr$\to $Ar &
      Pr$\to $Cl &
      Pr$\to $Re &
      Re$\to $Ar &
      Re$\to $Cl &
      Re$\to $Pr \\ 
    
    \hline 
    
    \multicolumn{2}{c|}{DER}      & \multirow{8}{*}{\rotatebox[origin=c]{90}{\textbf{TIL}}} & \underline{2.26} & \underline{2.38} & 2.93 & 2.29 & 3.37 & 3.09 & 2.31 & 2.62 & 3.18 & 3.75 & 1.77 & 1.72 \\
    \multicolumn{2}{c|}{DER++}    & & 2.25 & 2.40 & 2.77 & 2.60 & 3.34 & 3.26 & 2.32 & 2.51 & 3.12 & \underline{4.02} & \underline{3.41} & \underline{3.56} \\
    \multicolumn{2}{c|}{HAL}      & & 2.08 & 2.10 & 2.53 & 2.36 & 2.84 & 2.83 & 2.07 & 2.42 & 2.55 & 3.43 & 3.07 & 3.40 \\
    \multicolumn{2}{c|}{MSL}      & & \underline{2.26} & 2.35 & \underline{4.01} & \underline{4.02} & \underline{3.94} & \underline{3.65} & \underline{3.34} & \underline{3.32} & \underline{3.72} & \underline{4.02} & 2.24 & 2.14 \\
    \cline{0-0} \cline{4-15}
    \multicolumn{2}{c|}{CDTrans-S} & & 1.50 & 1.80 & 1.50 & 1.50 & 1.80 & 1.50 & 1.50 & 1.50 & 1.50 & 1.50 & 1.50 & 1.80 \\
    \multicolumn{2}{c|}{CDTrans-B} & & 1.00 & 1.20 & 1.40 & 1.40 & 1.20 & 1.40 & 1.40 & 1.00 & 1.65 & 1.70 & 1.12 & 1.20 \\
    \cline{0-0} \cline{4-15}
    \multicolumn{2}{c|}{Ours (ACC)} & & \textbf{24.44} & \textbf{25.18} & \textbf{26.20} & \textbf{21.25} & \textbf{26.64} & \textbf{23.54} & \textbf{22.89} & \textbf{24.21} & \textbf{29.44} & \textbf{26.25} & \textbf{26.27} & \textbf{31.25} \\
    \multicolumn{2}{c|}{Ours (FGT)} & & 7.33 & 8.17 & 10.97 & 10.90 & 9.93 & 11.06 & 9.05 & 10.32 & 13.89 & 12.75 & 11.05 & 19.17 \\
    
    \hline
    \hline
    
    \multicolumn{2}{c|}{DER} & \multirow{6}{*}{\rotatebox[origin=c]{90}{\textbf{CIL}}} & \textbf{4.40} & \underline{4.75} & \textbf{8.45} & 4.86 & \underline{10.63} & 9.30 & \underline{4.45} & \textbf{6.58} & \textbf{9.55} & 5.00 & 2.27 & 2.21 \\
    \multicolumn{2}{c|}{DER++} & & 3.89 & 4.93 & \underline{6.47} & \underline{5.52} & \textbf{11.11} & \textbf{11.06} & \textbf{4.49} & 5.50 & \underline{8.45} & \textbf{6.84} & \underline{4.00} & \underline{5.23} \\
    \multicolumn{2}{c|}{HAL} & & 4.01 & 3.27 & 6.24 & 4.70 & 7.16 & \underline{7.28} & 3.38 & \underline{5.18} & 4.66 & 3.54 & 3.03 & 4.74 \\
    \multicolumn{2}{c|}{MSL} & & 4.02 & 4.35 & 6.02 & 4.70 & 6.44 & 6.82 & 4.40 & 5.10 & 7.12 & 4.42 & 3.64 & 3.66 \\
    \cline{0-0} \cline{4-15}
    \multicolumn{2}{c|}{Ours (ACC)} & & \underline{4.15} & \textbf{4.77} & 6.01 & \textbf{5.56} & 8.32 & 6.35 & 3.92 & 5.02 & 6.29 & \underline{6.45} & \textbf{6.26} & \textbf{8.35} \\
    \multicolumn{2}{c|}{Ours (FGT)} & & 10.36 & 8.15 & 13.52 & 10.15 & 12.76 & 11.78 & 11.96 & 11.48 & 21.80 & 19.45 & 13.06 & 27.14 \\

    \hline
    \hline

    \multicolumn{3}{c|}{TVT (Static UDA)} & 74.89 & 86.82 & 89.47 & 82.78 & 87.95 & 88.27 & 79.81 & 71.94 & 90.13 & 85.46 & 74.62 & 90.56 \\

    \bottomrule
    
    \end{tabularx}
\vspace{1mm}
\caption{Comparison with SoTA methods' ACC on Office-Home. The best result in each experiment is highlighted with bold font, while the second-best is highlighted with underline font.}
\label{tab:officehome}
\vspace{-3mm}
\end{table*}

\begin{table*}[!]
\setlength\tabcolsep{1.0pt}
    \begin{tabularx}{\linewidth}{Y|Y|YYYYYY|Y|Y|YYYYYY}
    \hline
    \multicolumn{2}{c|}{DER} & clp & inf & pnt & qdr & rel & skt & \multicolumn{2}{c|}{DER} & clp & inf & pnt & qdr & rel & skt \\ 
    \hline
    
    \multirow{5}{*}{\rotatebox[origin=c]{90}{\textbf{TIL}}}
    & clp & -    & 0.53 & 0.55 & 0.49 & 0.51 & 0.53 & \multirow{5}{*}{\rotatebox[origin=c]{90}{\textbf{CIL}}} & clp & -   & 1.29 & 1.32 & \textbf{1.21} & 1.32 & 1.31 \\
    & inf & 0.55 & -    & 0.56 & 0.52 & 0.52 & 0.51 & & inf & 1.26 & -    & \textbf{1.32} & 1.32 & 1.24 & \textbf{1.32} \\
    & pnt & 0.54 & 0.52 & -    & 0.54 & 0.49 & 0.50 & & pnt & 1.32 & 1.33 & -    & \textbf{1.27} & 1.31 & \textbf{1.26} \\
    & qdr & 0.53 & 0.52 & 0.51 & -    & 0.49 & 0.52 & & qdr & 1.26 & \textbf{1.30} & 1.29 & -    & 1.19 & 1.26 \\
    & rel & 0.53 & 0.53 & 0.50 & 0.49 & -    & 0.51 & & rel & 1.29 & 1.27 & 1.22 & 1.23 & -    & 1.23 \\
    & skt & 0.51 & 0.52 & 0.51 & 0.52 & 0.53  & -   & & skt & 1.32 & 1.29 & 1.20 & \textbf{1.30} & 1.32 & -   \\

    \hline
    \multicolumn{2}{c|}{DER++} & clp & inf & pnt & qdr & rel & skt & \multicolumn{2}{c|}{DER++} & clp & inf & pnt & qdr & rel & skt \\ 
    \hline
    
    \multirow{5}{*}{\rotatebox[origin=c]{90}{\textbf{TIL}}}
    & clp & -    & 0.55 & 0.55 & 0.48 & 0.56 & 0.57 & \multirow{5}{*}{\rotatebox[origin=c]{90}{\textbf{CIL}}} & clp & -   & 1.30 & \textbf{1.34} & 1.18 & 1.24 & 1.27 \\
    & inf & 0.55 & -    & 0.53 & 0.55 & 0.52 & 0.52 & & inf & 1.23 & -    & 1.31 & \textbf{1.34} & 1.28 & 1.28 \\
    & pnt & 0.52 & 0.52 & -    & 0.56 & 0.57 & 0.54 & & pnt & 1.35 & 1.30 & -    & 1.24 & 1.26 & 1.23 \\
    & qdr & 0.57 & 0.58 & 0.57 & -    & 0.54 & 0.52 & & qdr & 1.32 & 1.27 & \textbf{1.35} & -    & \textbf{1.25} & \textbf{1.35} \\
    & rel & 0.55 & 0.54 & 0.54 & 0.55 & -    & 0.56 & & rel & 1.36 & 1.28 & 1.25 & \textbf{1.31} & - & \textbf{1.24} \\
    & skt & 0.55 & 0.52 & 0.54 & 0.57 & 0.53 & -    & & skt & 1.29 & 1.29 & 1.20 & 1.22 & 1.24 & -   \\

    \hline
    \multicolumn{2}{c|}{HAL} & clp & inf & pnt & qdr & rel & skt & \multicolumn{2}{c|}{HAL} & clp & inf & pnt & qdr & rel & skt \\ 
    \hline
    
    \multirow{5}{*}{\rotatebox[origin=c]{90}{\textbf{TIL}}}
    & clp & -    & 0.48 & 0.48 & 0.43 & 0.48 & 0.47 & \multirow{5}{*}{\rotatebox[origin=c]{90}{\textbf{CIL}}} & clp & -   & 0.88 & 0.87 & 0.81 & 0.90 & 0.86 \\
    & inf & 0.48 & -    & 0.50 & 0.45 & 0.49 & 0.45 & & inf & 0.91 & -    & 0.87 & 0.85 & 0.85 & 0.85 \\
    & pnt & 0.45 & 0.47 & -    & 0.51 & 0.49 & 0.48 & & pnt & 0.92 & 0.86 & -    & 0.84 & 0.86 & 0.86 \\
    & qdr & 0.48 & 0.46 & 0.46 & -    & 0.48 & 0.46 & & qdr & 0.88 & 0.91 & 0.88 & -    & 0.84 & 0.85 \\
    & rel & 0.47 & 0.45 & 0.45 & 0.48 & -    & 0.47 & & rel & 0.87 & 0.85 & 0.88 & 0.87 & -   & 0.92 \\
    & skt & 0.46 & 0.46 & 0.46 & 0.49 & 0.45 & -    & & skt & 0.94 & 0.91 & 0.89 & 0.90 & 0.88 & -   \\

    \hline
    \multicolumn{2}{c|}{MSL} & clp & inf & pnt & qdr & rel & skt & 
    \multicolumn{2}{c|}{MSL} & clp & inf & pnt & qdr & rel & skt \\ 
    \hline
    
    \multirow{5}{*}{\rotatebox[origin=c]{90}{\textbf{TIL}}}
    & clp & -    & 0.53 & 0.53 & 0.48 & 0.53 & 0.53 & \multirow{5}{*}{\rotatebox[origin=c]{90}{\textbf{CIL}}} & clp & -   & 1.30 & 1.32 & 1.20 & 1.30 & 1.30 \\
    & inf & 0.54 & -    & 0.55 & 0.54 & 0.54 & 0.50 & & inf & 1.24 & -    & 1.30 & 1.30 & 1.26 & 1.30 \\
    & pnt & 0.55 & 0.52 & -    & 0.54 & 0.52 & 0.52 & & pnt & 1.35 & 1.30 & -    & 1.26 & 1.30 & \textbf{1.26} \\
    & qdr & 0.55 & 0.55 & 0.55 & -    & 0.50 & 0.52 & & qdr & 1.30 & 1.30 & 1.30 & -    & 1.20 & 1.24 \\
    & rel & 0.55 & 0.53 & 0.52 & 0.53 & -    & 0.55 & & rel & 1.32 & 1.27 & 1.23 & 1.27 & -    & 1.23 \\
    & skt & 0.53 & 0.52 & 0.53 & 0.55 & 0.53  & -   & & skt & 1.30 & 1.29 & 1.20 & 1.25 & 1.30 & -   \\

    \hline
    \multicolumn{2}{c|}{CDTrans-S} & clp & inf & pnt & qdr & rel & skt & \multicolumn{2}{c|}{CDTrans-B} & clp & inf & pnt & qdr & rel & skt \\ 
    \hline
    
    \multirow{5}{*}{\rotatebox[origin=c]{90}{\textbf{TIL}}}
    & clp & -   & 0.00 & 0.00 & 0.00 & 0.00 & 0.00 & \multirow{5}{*}{\rotatebox[origin=c]{90}{\textbf{TIL}}} & clp & -   & 0.00 & 0.00 & 0.00 & 0.00 & 0.00 \\
    & inf & 0.30 & -   & 0.40 & 0.30 & 0.30 & 0.30 & & inf & 0.20 & -   & 0.20 & 0.30 & 0.20 & 0.20 \\
    & pnt & 0.20 & 0.30 & -   & 0.30 & 0.0 & 0.40 & & pnt & 0.20 & 0.40 & -   & 0.30 & 0.20 & 0.20 \\
    & qdr & 0.20 & 0.30 & 0.40 & -   & 0.30 & 0.30 & & qdr & 0.20 & 0.40 & 0.20 & -   & 0.20 & 0.20 \\
    & rel & 0.00 & 0.00 & 0.00 & 0.00 & -   & 0.00 & & rel & 0.00 & 0.00 & 0.00 & 0.00 & -   & 0.00 \\
    & skt & 0.30 & 0.30 & 0.40 & 0.30 & 0.30 & -   & & skt & 0.20 & 0.40 & 0.20 & 0.30 & 0.20 & -   \\

    \hline
    \multicolumn{2}{c|}{Ours (ACC)} & clp & inf & pnt & qdr & rel & skt & \multicolumn{2}{c|}{Ours (ACC)} & clp & inf & pnt & qdr & rel & skt \\ 
    \hline
    
    \multirow{5}{*}{\rotatebox[origin=c]{90}{\textbf{TIL}}}
    & clp & -   & \textbf{10.86} & \textbf{5.70} & \textbf{4.68} & \textbf{7.48} & \textbf{3.40} & \multirow{5}{*}{\rotatebox[origin=c]{90}{\textbf{CIL}}} & clp & -   & \textbf{1.33} & 0.63 & 0.45 & \textbf{2.06} & \textbf{2.17} \\
    & inf & \textbf{21.10} & -   & \textbf{10.12} & \textbf{2.03} & \textbf{13.50} & \textbf{8.77} & & inf & \textbf{2.59} & -   & 1.26 & 0.25 & \textbf{1.64} & 1.02 \\
    & pnt & \textbf{22.80} & \textbf{10.84} & -   & \textbf{2.52} & \textbf{14.61} & \textbf{9.96} & & pnt & \textbf{2.98} & \textbf{1.31} & - & 0.28 & \textbf{1.77} & 1.18 \\
    & qdr & \textbf{16.43} & \textbf{2.32} & \textbf{3.85} & -   & \textbf{6.96} & \textbf{4.64} & & qdr & \textbf{1.94} & 0.29 & 0.48 & -   & 0.89 & 0.58 \\
    & rel & \textbf{26.22} & \textbf{12.26} & \textbf{11.85} & \textbf{2.57} & -   & \textbf{10.23} & & rel & \textbf{3.24} & \textbf{1.50} & \textbf{1.49} & 0.31 & -   & 1.20 \\
    & skt & \textbf{27.62} & \textbf{11.58} & \textbf{11.07} & \textbf{5.15} & \textbf{13.93} & -   & & skt & \textbf{3.49} & \textbf{1.35} & \textbf{1.31} & 0.58 & \textbf{1.64} & -   \\

    \hline
    \multicolumn{2}{c|}{Ours (FGT)} & clp & inf & pnt & qdr & rel & skt & \multicolumn{2}{c|}{Ours (FGT)} & clp & inf & pnt & qdr & rel & skt \\ 
    \hline
    
    \multirow{5}{*}{\rotatebox[origin=c]{90}{\textbf{TIL}}}
    & clp & -   & 4.22 & 6.42 & 1.44 & 10.51 & 2.44 & \multirow{5}{*}{\rotatebox[origin=c]{90}{\textbf{CIL}}} & clp & -   & 5.19 & 8.12 & 1.25 & 11.39 & 2.54 \\
    & inf & 7.73 & -   & 3.93 & 0.81 & 4.96 & 3.45 & & inf & 9.46 & -   & 4.59 & 0.96 & 6.27 & 4.07 \\
    & pnt & 9.04 & 4.21 & -   & 0.97 & 5.42 & 3.88 & & pnt & 11.18 & 5.22 & -   & 1.17 & 6.73 & 4.61 \\
    & qdr & 6.19 & 0.89 & 1.50 & -   & 2.83 & 1.91 & & qdr & 7.37 & 1.08 & 1.83 & -   & 3.41 & 2.32 \\
    & rel & 10.25 & 4.70 & 4.59 & 1.06 & -   & 3.60 & & rel & 12.36 & 5.89 & 5.79 & 1.24 & -   & 4.49 \\
    & skt & 11.14 & 4.32 & 4.38 & 1.85 & 5.52 & -   & & skt & 13.20 & 5.17 & 5.51 & 2.47 & 6.70 & -   \\

    \hline
    \multicolumn{2}{c|}{TVT (Static UDA)} & clp & inf & pnt & qdr & rel & skt \\ 
    \hline
    
    \multirow{5}{*}{\rotatebox[origin=c]{90}{\textbf{Static}}}
    & clp & -  & 24.2 & 48.9 & 15.5 & 63.9 & 50.7 \\
    & inf & 43.5 & - & 44.9 & 6.5 & 58.8 & 37.6 \\
    & pnt & 52.8 & 23.3 & - & 6.6 & 64.6 & 44.5 \\
    & qdr & 31.8 & 6.1 & 15.6 & - & 23.4 & 18.9 \\
    & rel & 58.9 & 26.3 & 56.7 & 9.1 & -   & 45.0 \\
    & skt & 60.0 & 21.1 & 48.4 & 16.6 & 61.7 & -   \\

    \bottomrule
    \end{tabularx}
\vspace{1mm}
\caption{Comparison with SoTA methods' ACC on DomainNet. The rows represent the source dataset, while the columns represent the target dataset. The best result in each experiment is highlighted with bold font.}
\label{tab:domainnet-sota-2}
\vspace{-3mm}
\end{table*}

We conducted a comprehensive analysis of all baselines' capabilities in mitigating {\it feature alignment catastrophic forgetting}. The results for both the Task-Incremental Learning (TIL) and Class-Incremental Learning (CIL) scenarios are presented in Tables \ref{tab:office31}, \ref{tab:officehome}, and \ref{tab:domainnet-sota-2}. These tables compare the average accuracy of all baselines alongside CDCL's additional coverage of average forgetting.

Our findings reveal that CDCL exhibits superior performance over all continual baselines in the TIL setup. This success can be attributed to CDCL's {\it inter- intra-task cross-attention} mechanism, which effectively retains previous task information in the attention maps (queries and values projections). Additionally, CDCL maximizes intra-task knowledge in the key vectors from both source and target domains. Furthermore, CDCL excels in scenarios where the target domain is closely related to the source domain, such as in the MNIST$\leftrightarrow$USPS experiments. This success is primarily due to CDCL's {\it intra-task center-aware pseudo-labeling} mechanism, which enables successful semi-supervised prediction of the target domain. Remarkably, CDCL's performance in the D$\rightarrow$W and W$\rightarrow$D experiments further reinforces this claim. Significantly, in the DomainNet experiments, CDCL stands out as the only continual method capable of demonstrating any signal of learning and catastrophic forgetting mitigation.

\begin{figure}
    \centering
    \includegraphics[width=\columnwidth]{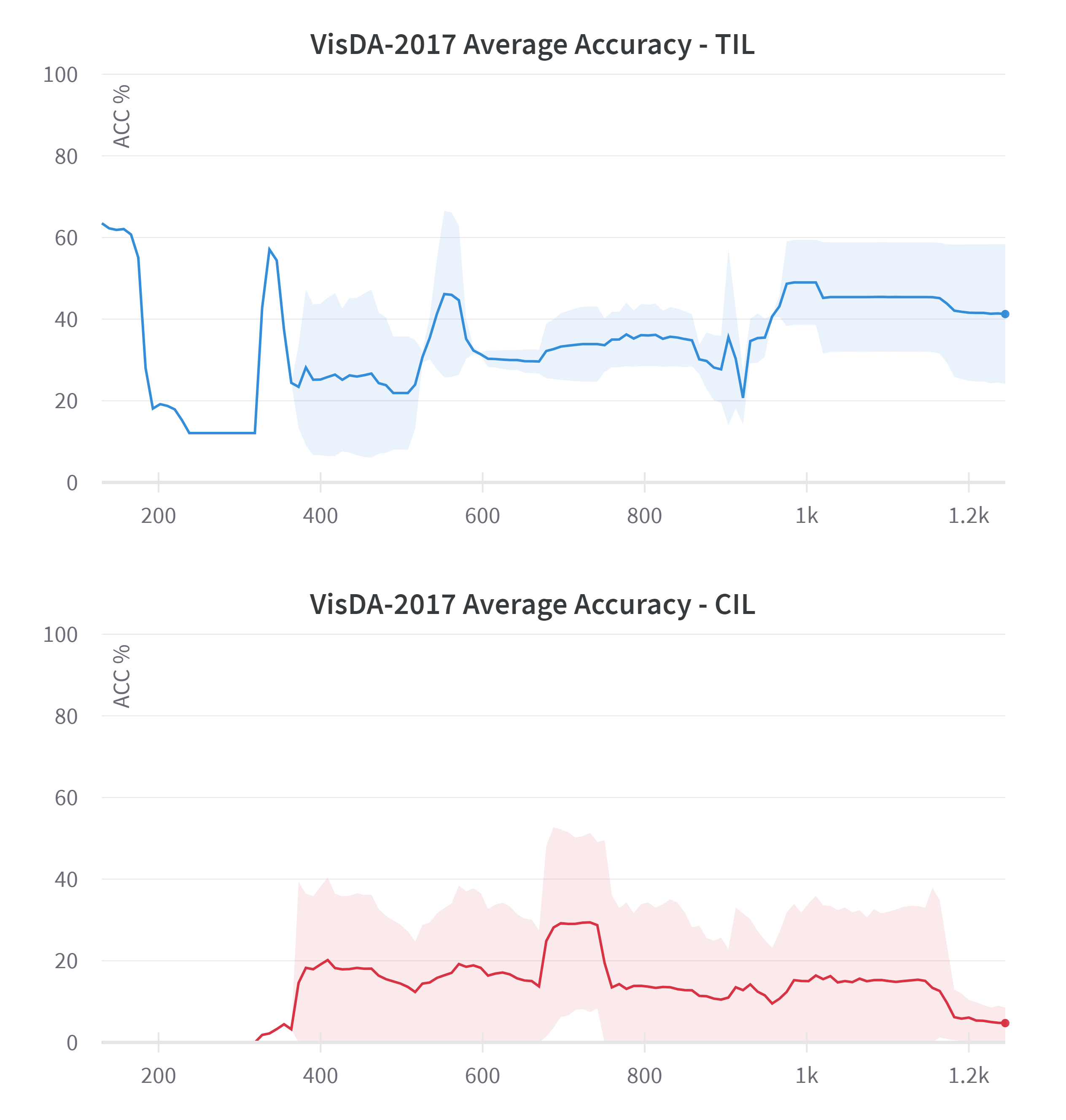}
    \caption{The evolution of CDCL's ACC in the VisDA-2017 for both TIL and CIL scenarios. The shared area represents the standard deviation of $R_{i,j}, i \in [1, j]$, the accuracy on such a task by a model that learned only previous tasks.}
    \label{fig:visda}
\end{figure}

However, it is important to acknowledge certain limitations. First, all continual learning that presents task-related parameters, such as CDCL, suffers from a theoretical infinitely growing numbers of parameter. Usually, this weakness is overlooked because of the small footprint added to the computation complexity of the model as more tasks are introduced. Additionally, such limitation is easy handled by spliting the parameters records into the primary memory (RAM) and the secondary memories (HDDs or SSDs), loading only the necessary parameters during inference time. Finally, all methods, including CDCL, still need improvements in addressing the challenges of the CIL scenario. In tackling such demanding experiments, CDCL adopts the experimental replay strategy introduced by DER and DER++ to recover forgotten information swiftly. Consequently, all three baselines perform similarly in these scenarios. The hypothesis behind this method lies in the fact that forgotten knowledge may still reside in residual forms within the network parameters, which can be partially recovered through experience replay. For future work, it is essential to devise an improved mechanism that preserves inter-task common knowledge while effectively adapting to new tasks, problem also depict on Fig. \ref{fig:visda} with the VisDA-2017 benchmark. Addressing this aspect will contribute to enhancing CDCL's performance in handling the CIL scenario and advancing the field of cross-domain continual learning.\footnote{The work presented on \cite{TwoStreamCL} seems to tackle the CIL scenario. However, the authors didn't release their method source code; and their experimental setup is very different from the standard continual learning procedure, presented here.}

\subsection{Ablation study}
\label{subsec:ablation-study}

In our investigation, we delved into the influence of the three losses-block on the final ACC performance of CDCL. Additionally, we explored the impact of the {\it inter- intra-task cross-attention}, which involves evaluating how a standard simple attention mechanism on the source domain enables the network to generalize to the target domain. The results of our analysis are presented in Table \ref{tab:ablation}.

The examination of the losses-block revealed their relative importance in determining CDCL's performance. Notably, the intra-task loss emerged as the most critical factor, followed closely by the rehearsal loss. This finding suggests that leveraging intra-task information plays a significant role in enhancing the model's performance in CDCL.

Interestingly, the omission of the inter-task loss had a relatively low-performance impact, indicating that much of the knowledge is effectively preserved in the $\mathbf{K}_i$ and $\mathbf{b}_i$ projections. This observation highlights the effectiveness of the intra-task and rehearsal losses in capturing and retaining crucial information during the learning process.

Furthermore, in the context of the class-incremental learning (CIL) scenario, the rehearsal loss demonstrated a higher level of importance in preserving and adapting inter-task information. This aspect is especially valuable when addressing the continual learning of new classes, as the rehearsal loss aids in retaining knowledge from previous tasks, facilitating effective adaptation to new ones.

Finally, it was noted that when utilizing the standard simple attention mechanism, CDCL's main contribution is compromised, leading to a performance that closely resembles that of DER (Domain-Adversarial Network for Domain-Adaptive Visual Recognition) and DER++. This outcome underscores the unique and significant impact of CDCL's approach to cross-domain continual learning and the importance of its inter-intra-task cross-attention mechanism in achieving superior performance over alternative methods.

\begin{table}[t]
\setlength\tabcolsep{1.3pt}
\begin{center}
    \resizebox{\columnwidth}{!}{
    \begin{tabular}{c|ccc|cc|cc}
    \hline
    \multirow{2}{*}{Experiment} 
        &  
            \multicolumn{3}{c|}{Loss / Module} & \multicolumn{2}{c|}{MN$\to$US} &  \multicolumn{2}{c}{US$\to$MN}\\ 
        & 
            \textbf{$\mathcal{L}^\text{CIL}$} &
            \textbf{$\mathcal{L}^\text{TIL}$} &
            \textbf{$\mathcal{L}_\mathrm{R}$} &
            \textbf{TIL} &
            \textbf{CIL} & 
            \textbf{TIL} &
            \textbf{CIL} \\

    \hline

    - & \checkmark & \checkmark & \checkmark & 91.91 & 66.73 & 81.48 & 52.50 \\
    A & -          & \checkmark & \checkmark & 81.88 & 63.71 & 67.31 & 47.86  \\
    B & \checkmark &    -       & \checkmark & 59.17 & 46.33 & 60.20 & 34.65 \\
    C & \checkmark & \checkmark &   -        & 68.71 & 19.59 & 71.72 & 15.83 \\

    \hline
    \hline

    Simple attention & \checkmark* & \checkmark* & \checkmark* & 62.72 & 29.82 & 73.20 & 29.50 \\

    \bottomrule
    \end{tabular}}
\end{center}
\vspace{-2mm}
\caption{Ablation study analyzing the effect of each loss module into CDCL's ACC.}
\vspace{-3mm}
\label{tab:ablation}
\end{table}

\section{Conclusion}
\label{sec:conclusion}

In this paper, we introduced a novel framework called Cross-Domain Continual Learning (CDCL), which effectively tackles the challenge of unsupervised cross-domain problems in the task-incremental learning setup. CDCL capitalizes on two key mechanisms: the inter-intra-task cross-attention block and the intra-task center-aware pseudo-labeling procedure. Leveraging the transformer architecture, CDCL enables simultaneous learning across multiple domains in a continual learning context.

Through extensive numerical experiments, we demonstrated the superiority of CDCL over its baselines in the task-incremental learning scenario. The results underscore the effectiveness of the inter-intra-task cross-attention and the center-aware pseudo-labeling strategies in enhancing CDCL's performance.

Looking ahead, our future work will focus on further refining CDCL in the direction of fully class-incremental learning. We aim to extend the capabilities of CDCL to effectively handle the continual introduction of new classes, thus advancing the field of cross-domain continual learning and its practical applications.

{\small
\bibliographystyle{ieee_fullname}
\bibliography{egbib}
}

\end{document}